\documentclass[a4paper, 10pt, conference]{ieeeconf}      % Use this line for a4
\usepackage{FG2025}
\usepackage{amsmath}
\usepackage{graphicx}
\usepackage[utf8]{inputenc}
\usepackage{cite}  
\usepackage{hyperref}  % Enlaces clicables en PDF
\usepackage{stfloats}
\usepackage{booktabs}

\FGfinalcopy % *** Uncomment this line for the final submission

\IEEEoverridecommandlockouts                              % This command is only
\def\FGPaperID{****} % *** Enter the FG2025 Paper ID here

\title{\LARGE \bf
TrustSkin: A Fairness Pipeline for Trustworthy Facial Affect Analysis Across Skin Tone}

\author{\parbox{16cm}{\centering
    {\large Ana M. Cabanas\textsuperscript{1,*}, Alma Pedro\textsuperscript{2}, Domingo Mery\textsuperscript{2}}\\
    {\normalsize
    \textsuperscript{1}Departamento de Física, Universidad de Tarapacá, Arica, 1010069, Chile\\
    \textsuperscript{2}Departamento de Ciencia de la Computación, Facultad de Ingeniería, Pontificia Universidad Católica de Chile, Santiago, 7820436, Chile\\
    \textsuperscript{*}Corresponding author: Ana M. Cabanas, \texttt{acabanas@academicos.uta.cl}}
    }
}

\usepackage{fancyhdr}
\begin{document}

\ifFGfinal
\thispagestyle{empty}
\pagestyle{empty}
\else
\author{Anonymous FG2025 submission\\ Paper ID \FGPaperID \\}
\pagestyle{plain}
\fi
\maketitle

\thispagestyle{fancy}

\begin{abstract}
Understanding how facial affect analysis (FAA) systems perform across different demographic groups requires reliable measurement of sensitive attributes such as ancestry, often approximated by skin tone, which itself is highly influenced by lighting conditions. This study compares two objective skin tone classification methods: the widely used Individual Typology Angle (ITA) and a perceptually grounded alternative based on Lightness ($L^*$) and Hue ($H^*$). Using AffectNet and a MobileNet-based model, we assess fairness across skin tone groups defined by each method. Results reveal a severe underrepresentation of dark skin tones ($\sim$2$\%$), alongside fairness disparities in F1-score (up to 0.08) and TPR (up to 0.11) across groups. While ITA shows limitations due to its sensitivity to lighting, the \(H^*\)-\(L^*\) method yields more consistent subgrouping and enables clearer diagnostics through metrics such as Equal Opportunity. Grad-CAM analysis further highlights differences in model attention patterns by skin tone, suggesting variation in feature encoding. To support future mitigation efforts, we also propose a modular fairness-aware pipeline that integrates perceptual skin tone estimation, model interpretability, and fairness evaluation. These findings emphasize the relevance of skin tone measurement choices in fairness assessment and suggest that ITA-based evaluations may overlook disparities affecting darker-skinned individuals.
\end{abstract}

%%%%%%%%%%%%%%%%%%%%%%%%%%%%%%%%%%%%%%%%%%%
\section{INTRODUCTION}

Predictive algorithms and biometric systems are increasingly used in critical areas such as healthcare, security, and human-computer interaction~\cite{Drozdowski2020}. However, these systems remain prone to bias arising from demographic imbalances in training data and algorithmic design flaws~\cite{Kolla, Schwartz2022, Drozdowski2020}. In computer vision applications like EmotionAI and Facial Affect Analysis (FAA), such biases often result in consistent performance disparities across attributes like age, sex, and skin tone~\cite{Mattioli2024, Xu2020, Rizbi2025}. Given the sensitive deployment of FAA in psychological evaluation, driver monitoring, and educational feedback~\cite{Drozdowski2020, Hosseini2025, Silberg2019}, ensuring fairness, transparency, and robustness across demographic groups is essential.

A key challenge in fairness research is the reliable measurement of sensitive attributes such as ancestry or skin tone, which is highly affected by lighting conditions. The \textit{Gender Shades} study exemplified this, showing error rates of 34.7$\%$ for darker-skinned women versus 0.8$\%$ for lighter-skinned men in commercial systems~\cite{Buolamwini2018}. These disparities raise concerns about identity construction in facial datasets, where categories like gender and ancestry are often treated as fixed despite lacking standardized definitions~\cite{Scheuerman2020, Rizbi2025, Xu2020}. Ancestry labels are socially constructed and inconsistently applied~\cite{Khan2021}, while skin tone annotations are influenced by lighting, sensor variation, and annotator bias~\cite{Schumann2023, Heldreth2024}. Widely used scales such as the Fitzpatrick Skin Type (FST)~\cite{Fitzpatrick1988} are biased toward lighter skin and unsuitable for fairness benchmarking~\cite{Buolamwini2018, Schumann2023, Scheuerman2020, Khan2021}. As an alternative, the Individual Typology Angle (ITA)~\cite{Wilkes2015, Merler2019, Karkkainen2021} is commonly used but presents critical limitations under real-world conditions~\cite{Thong2023, Schumann2023}.

Although skin tone bias has been extensively studied in face recognition systems~\cite{Breckon2023, Schwartz2022, Huebner, Drozdowski2020, Wang2021MetaBN}, it remains comparatively under-explored FAA~\cite{Hazirbas2022, Heldreth2024, Thong2023, Rhue2018}. While some fairness-aware strategies, such as objective tone classification and dataset rebalancing, have been proposed, their effectiveness in affective computing remains largely unvalidated~\cite{FAn2023, Hosseini2025}. 

Recent work has investigated fairness in FAA through diverse lenses, including algorithmic perception and trust~\cite{Lee2018}, counterfactual fairness~\cite{Cheong2022}, multimodal bias analysis~\cite{Schmitz2022}, domain adaptation for demographic bias~\cite{Singhal2025}, and evaluation frameworks that go beyond accuracy~\cite{Fromberg2024}. However, despite these advances, none of these studies explicitly address the role of skin tone as a source of bias in affective systems.

Modeling skin tone under real-world conditions remains a persistent challenge, as illumination, sensor variation, and background context can introduce artifacts that distort both classification and subgroup attribution. Prior work in skin detection has shown that hybrid color space representations (e.g., RGB, HSV, YCbCr) offer more robust tone modeling than RGB alone~\cite{Oliveira2018, kolkur2017,cook2025}. This is particularly relevant to FAA, where widely used proxy metrics like the Individual Typology Angle (ITA) oversimplify skin tone and fail to account for lighting variability~\cite{Dehdashtian2024}. To address this, we propose a perceptually grounded measurement method based on Lightness ($L^*$) and Hue ($H^*$) to improve subgroup consistency and fairness evaluation under uncontrolled imaging conditions.

We empirically assess this approach using the AffectNet dataset~\cite{Mollahosseini2019} and a MobileNet-based classifier, comparing fairness metrics across skin tone groups defined by both ITA and our $L^*$–$H^*$ method. Our analysis reveals a severe demographic imbalance in the dataset, with less than 2\% of samples representing the darkest skin tone group—echoing similar disparities observed in dermatology datasets~\cite{Corbin2023, Kinyanjui2019} and computer vision benchmarks~\cite{Buolamwini2018, Schumann2023}. Furthermore, we show that ITA-based groupings produce erratic and potentially misleading fairness metrics for darker-skinned individuals, likely due to illumination-related measurement artifacts—a finding consistent with prior work~\cite{kolkur2017, Corbin2023}. In contrast, the $L^*$–$H^*$ approach yields more stable subgroupings and fairness evaluations, though still limited by underrepresentation. Attempts to correct this imbalance through targeted data augmentation failed to improve fairness on the unbalanced test set, corroborating findings that post-hoc mitigation is ineffective when upstream bias persists~\cite{Dehdashtian2024}.

%%%%%%%%%%%%%%%%%%%%%%

\section{METHODOLOGY}
\label{sec:methodology}

To investigate the impact of skin tone measurement on fairness in FAA, we compared two objective skin tone estimation methods. Our pipeline involved dataset selection, preprocessing, skin pixel segmentation, color space conversion, tone stratification using ITA and Hue-Lightness ($H^*L^*$), and subsequent demographic analysis.

\subsection{Dataset and Preprocessing}
We used the AffectNet dataset~\cite{Mollahosseini2019}, a large-scale resource for facial affect analysis containing over one million facial images collected from the web. Approximately 450,000 of these images include manual annotations for categorical emotions (neutral, happy, sad, surprise, fear, disgust, anger, contempt). While AffectNet provides extensive variability in facial pose, lighting, and expression, it was not designed with demographic balance or fairness in mind, and includes a substantial number of grayscale or near-grayscale images. These characteristics limit its suitability for color-based skin tone analysis required in fairness evaluations.

We applied a preprocessing pipeline to address these limitations to filter out grayscale and low-color content images from the training and validation splits. Face regions were automatically detected and cropped using MTCNN to minimize background interference and improve skin tone estimation. The images were then converted to the YCrCb color space. Skin pixels were segmented using a combination of Otsu's thresholding and chrominance-based filtering, applied in both YCrCb and HSV spaces following validated thresholds from prior work~\cite{kolkur2017, Merler2019}. Finally, the average RGB values of the segmented skin regions were transformed into the CIE Lab color space for downstream analysis.

\subsection{Objective Skin Tone Estimation and Stratification}

To evaluate the role of skin tone in fairness analysis, we employed two objective classification methods. First, we used the widely adopted Individual Typology Angle (ITA) \cite{Wilkes2015,Merler2019,Bino2023,Kinyanjui2019}. Second, we implemented a perceptually grounded alternative based on Lightness ($L^*$) and Hue ($H^*$), designed to improve robustness to illumination variability and better capture subtle chromatic differences \cite{Thong2023,cook2025}. Both approaches were applied to the training and validation splits, enabling direct comparison of their effectiveness in capturing meaningful tone distinctions.

For ITA-based classification, we extracted color values from the CIE Lab color space and applied the standard formula:

\[
\text{ITA} = \arctan\left(\frac{L^* - 50}{b^*}\right) \cdot \frac{180}{\pi}.
\]

In this expression, $L^*$ denotes perceptual lightness and $b^*$ the blue–yellow chromaticity. ITA quantifies skin tone by evaluating the relationship between these two components: higher ITA values generally correspond to lighter skin tones, while lower values indicate darker complexions.

\begin{figure}[ht]
    \centering
    \includegraphics[width=\linewidth]{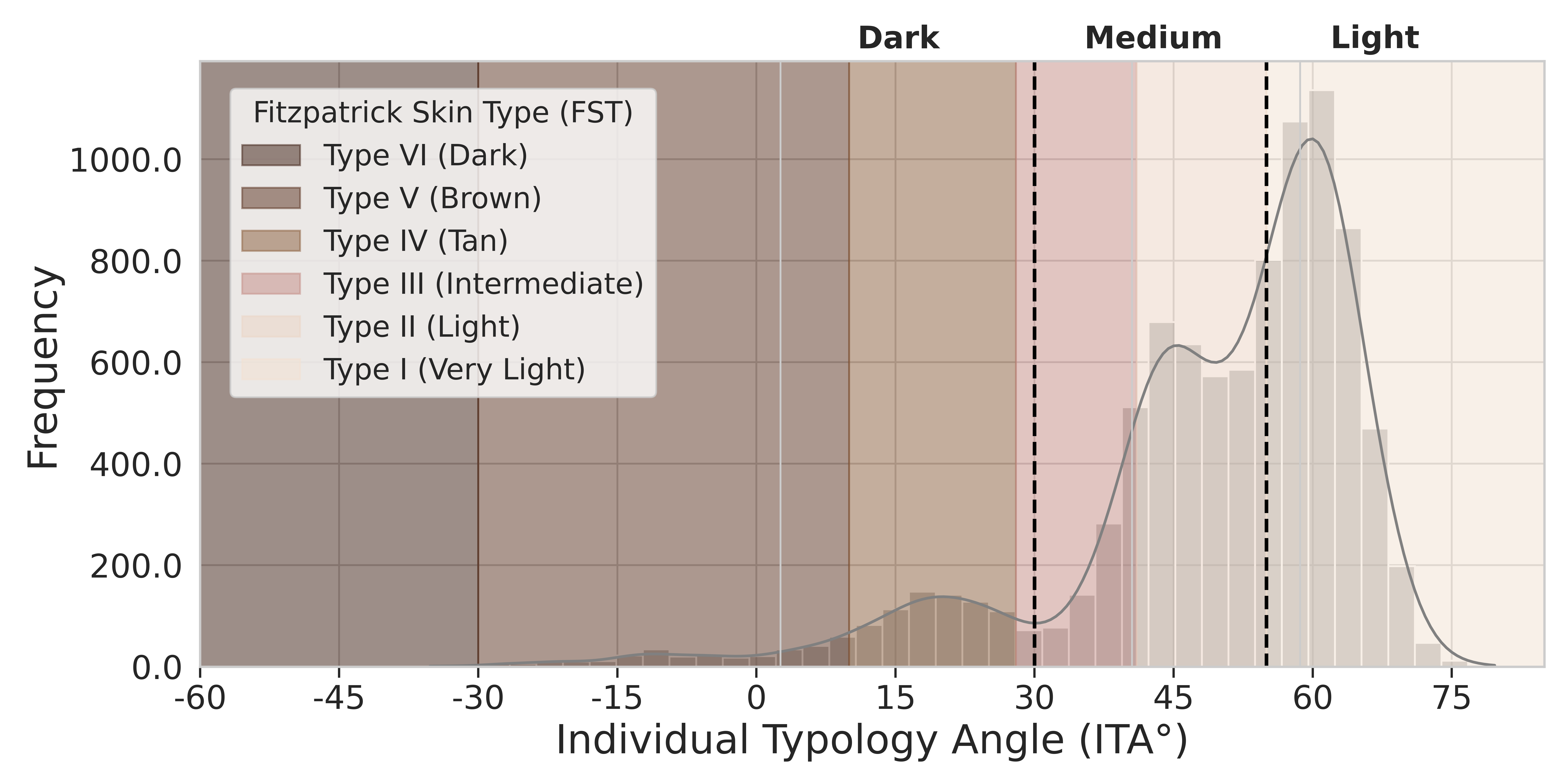}
    \caption{
    Distribution of ITA values in the cleaned AffectNet training set. Background shading reflects Fitzpatrick types; dashed lines mark our custom thresholds for Light (ITA $> 55^\circ$), Medium (30$^\circ$ $\leq$ ITA $\leq$ 55$^\circ$), and Dark (ITA $< 30^\circ$).}
    \label{fig1}
\end{figure}

To contextualize our ITA-based classification within established dermatological frameworks, we mapped ITA values to the Fitzpatrick Skin Type (FST) scale~\cite{Fitzpatrick1988}. Originally developed for dermatology to describe skin responses to ultraviolet radiation, the FST provides a coarse, numerical classification of human skin color. While not intended for computer vision or demographic analysis, it remains a commonly used reference \cite{Chiu2024,Heldreth2024, Schumann2023}. Figure~\ref{fig1} shows the distribution of ITA values in our cleaned AffectNet training subset, overlaid with approximate FST groupings. The distribution is clearly skewed toward lighter skin tones, with the majority of images falling into the Type I–III range (ITA $>$ 28°), indicating substantial overrepresentation of lighter complexions. In contrast, lower ITA intervals (FST Types V and VI) are sparsely populated, revealing a pronounced underrepresentation of darker skin tones.

For our analysis, we simplified the FST-based categorization into three broader skin tone groups—Light (ITA $> 55^\circ$), Medium ($30^\circ \leq$ ITA $\leq 55^\circ$), and Dark (ITA $< 30^\circ$), as indicated by the vertical dashed lines in Figure~\ref{fig1}. This grouping facilitates statistically meaningful comparisons while aligning with prior work on fairness evaluation \cite{Schumann2023,Kinyanjui2019a,Montoya2024}. However, this simplification reduces granularity by merging six FST levels into three, potentially overlooking finer tonal differences. It may also introduce classification ambiguity near group boundaries, where small measurement shifts due to lighting or noise can affect assignment. Nonetheless, discretizing a continuous and complex trait like skin tone is a common and pragmatic approach in fairness studies, serving as an operational compromise rather than a definitive categorization.

Despite these limitations, this simplified grouping allowed us to conduct a comparative fairness analysis while acknowledging that a more nuanced approach might be desirable with a larger, more balanced dataset. The severe underrepresentation of darker skin tones, even after this simplification (as seen in the small proportion of samples to the left of the Dark boundary in Figure~\ref{fig1}), highlights a critical challenge in fairness research and the need for careful consideration of measurement methods and dataset biases.

Prior studies have shown that ITA correlates with melanin index and offers a reproducible, colorimetric proxy for skin pigmentation~\cite{Heldreth2024, Thong2023, Merler2019}, reducing dependence on socially constructed and inconsistently annotated demographic labels such as perceived ancestry~\cite{Scheuerman2020, Khan2021}. Despite its practical advantages, ITA presents a notable methodological limitation: while it incorporates lightness ($L^*$) and blue–yellow chromaticity ($b^*$), it entirely omits the $a^*$ component that encodes the green–red axis. As a result, ITA fails to capture hue information, which plays a critical role in perceived skin color differences—particularly in populations with diverse undertones~\cite{Thong2023}.

To address the limitations of ITA, specifically its sensitivity to variations in illumination and its loss of chromatic information, we implemented an alternative skin tone classification method using ($L^*$) and Hue ($H^*$). Hue is derived from the CIE Lab color space using the following formula:

\[
\text{Hue} = \arctan\left(\frac{b^*}{a^*}\right).
\]

Unlike ITA, which omits the $a^*$ channel, the Hue-Lightness  (\(H^*\) - \(L^*\)) color space incorporates both axes, allowing finer differentiation of undertones, particularly important in distinguishing reddish versus yellowish skin hues~\cite{Thong2023}. To stratify skin tones, we initially applied thresholds to the $L^*$ dimension: Light: ($L^* > 67$), Medium: ($37 \leq L^* \leq 67$), and Dark: ($L^* < 37$). 

\begin{figure}[ht]
    \centering
    \includegraphics[width=\linewidth]{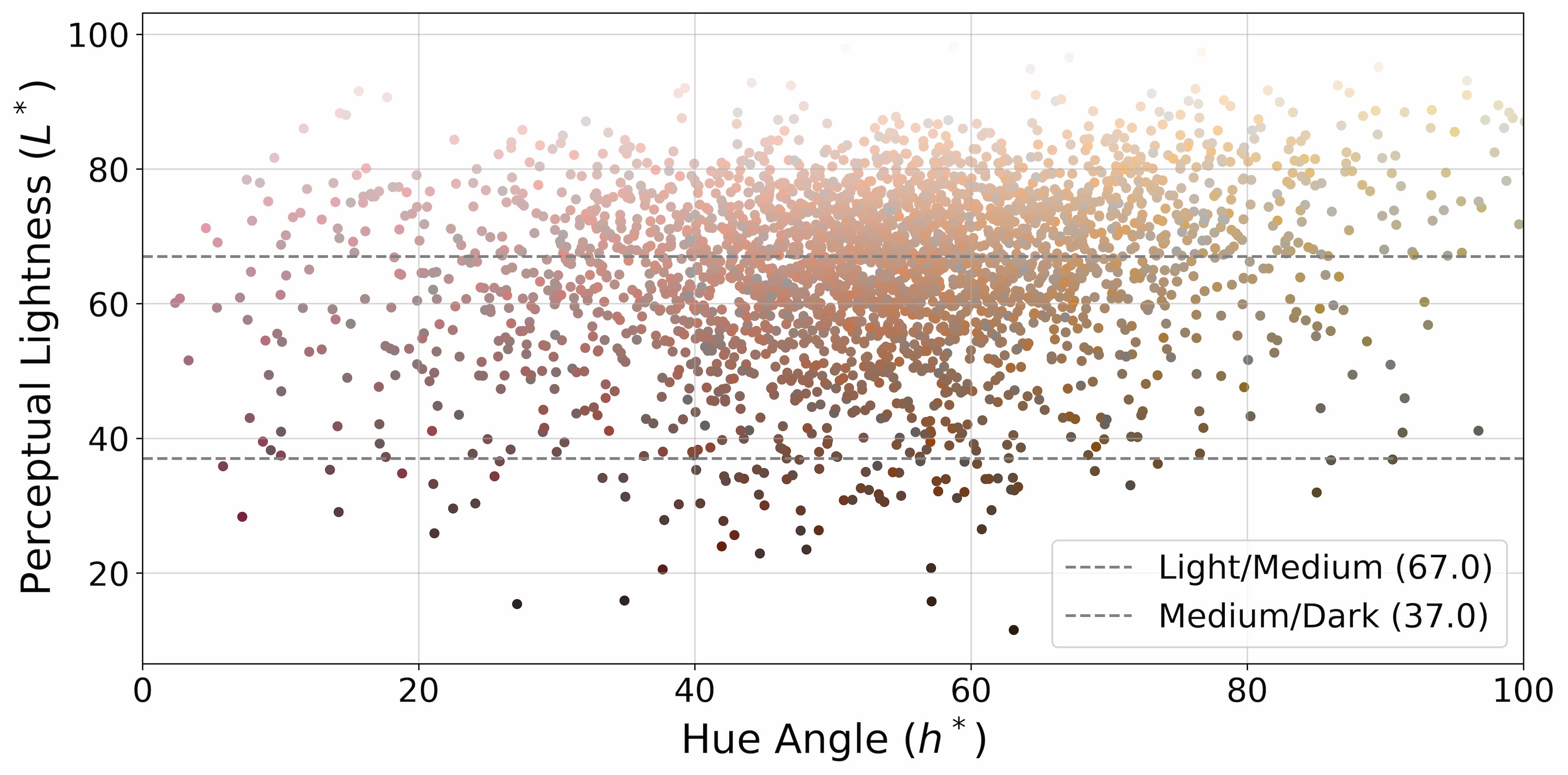}
    \caption{Skin tone distribution in the \((H^*\)–\(L^*)\) color space. Point colors correspond to their respective RGB values. Dashed horizontal lines denote \(L^*\) thresholds for skin tone classification: Light (\(L^* > 67.0\)), Medium (\(37.0 \leq L^* \leq 67.0\)), and Dark (\(L^* < 37.0\)).}
    \label{fig2}
\end{figure}

Figure~\ref{fig2} illustrates the distribution of estimated skin tones in the \(H^*\)-\(L^*\) space. The x-axis shows the Hue angle (\(H^*\)), capturing perceived undertones (e.g., red to yellow), while the y-axis represents lightness (\(L^*\)), from 0 (black) to 100 (white). Each point corresponds to the RGB-estimated skin tone of an image, with color reflecting its original RGB value. Dashed horizontal lines mark the \(L^*\) thresholds used to define Light, Medium, and Dark categories. The plot reveals clear clustering: lighter tones concentrate at higher \(L^*\) values, and darker tones at lower ones. However, significant variation along the \(H^*\) axis, even within the same \(L^*\) range, shows that the hue captures the tonal subtleties missed by lightness alone. This supports the use of both dimensions for more accurate skin tone classification.

Still, relying solely on \(L^*\) can misclassify darker skin tones with elevated lightness due to lighting or exposure. To address this, we apply a brown-tone override that reclassifies pixels as Dark based on RGB ranges and Hue, independently of their initial \(L^*\) label. Specifically, a pixel qualifies if: \(100 \leq R \leq 170\), \(60 \leq G \leq 110\), \(40 \leq B \leq 85\), with \(R > G > B\), and \((R - G) < 30\), \((G - B) < 25\). The Hue angle must fall within \(20^\circ \leq H^* \leq 50^\circ\), capturing typical brown hues. This rule improved classification in underrepresented groups, correctly reassigning two borderline cases (see last row of Figure~\ref{fig:hue_method}). While effective in this dataset, the override rule relies on empirical thresholds and may require tuning to generalize across other datasets.

\begin{figure}[htbp]
    \centering
    \includegraphics[width=\linewidth]{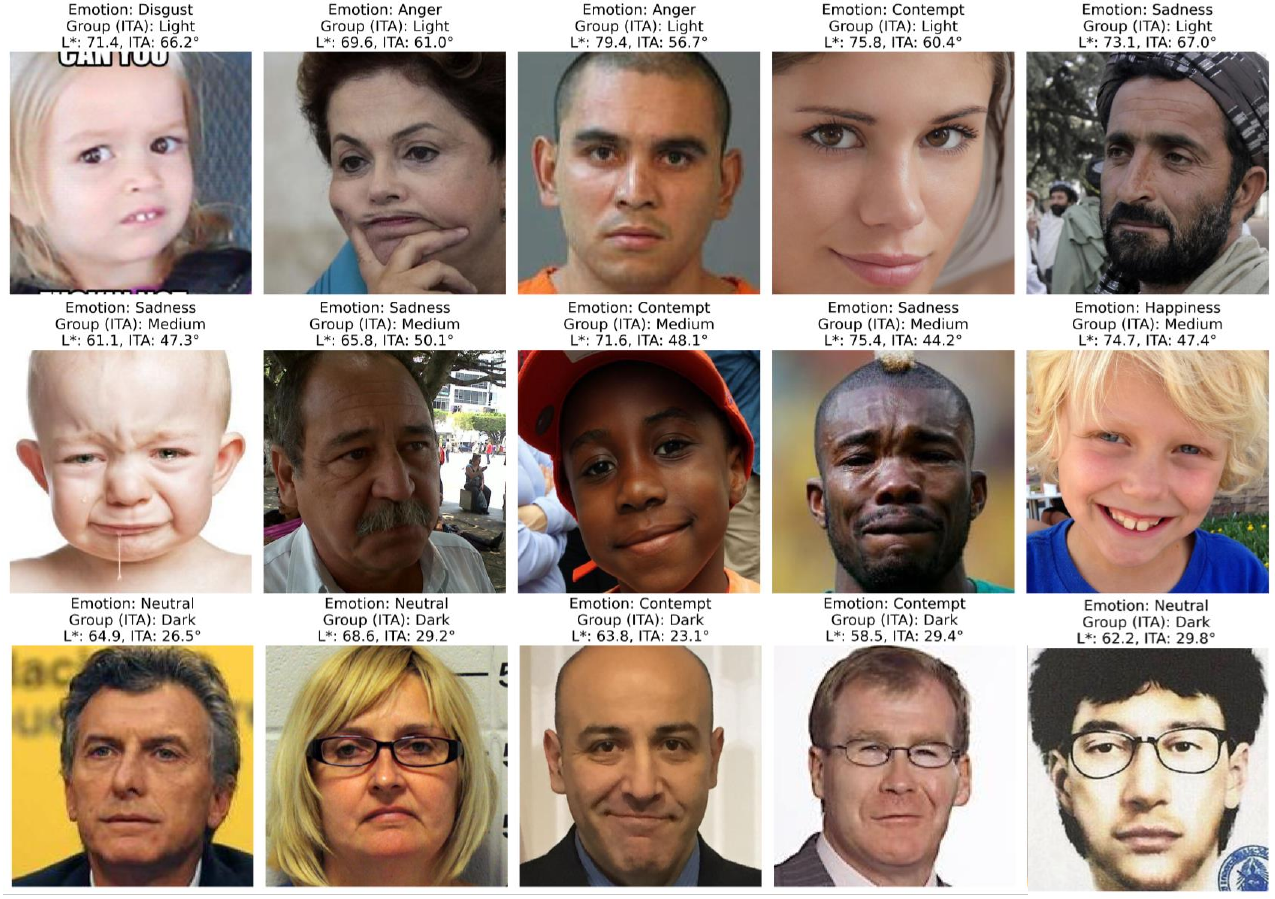}
    \caption{
    Skin tone classification using the Individual Typology Angle (ITA). Images were randomly sampled from the training subset of the AffectNet dataset. While commonly used, ITA is sensitive to illumination and often misclassifies Medium and Dark skin tones.}
    \label{fig:ita_method}
\end{figure}

\begin{figure}[htbp]
    \centering
    \includegraphics[width=\linewidth]{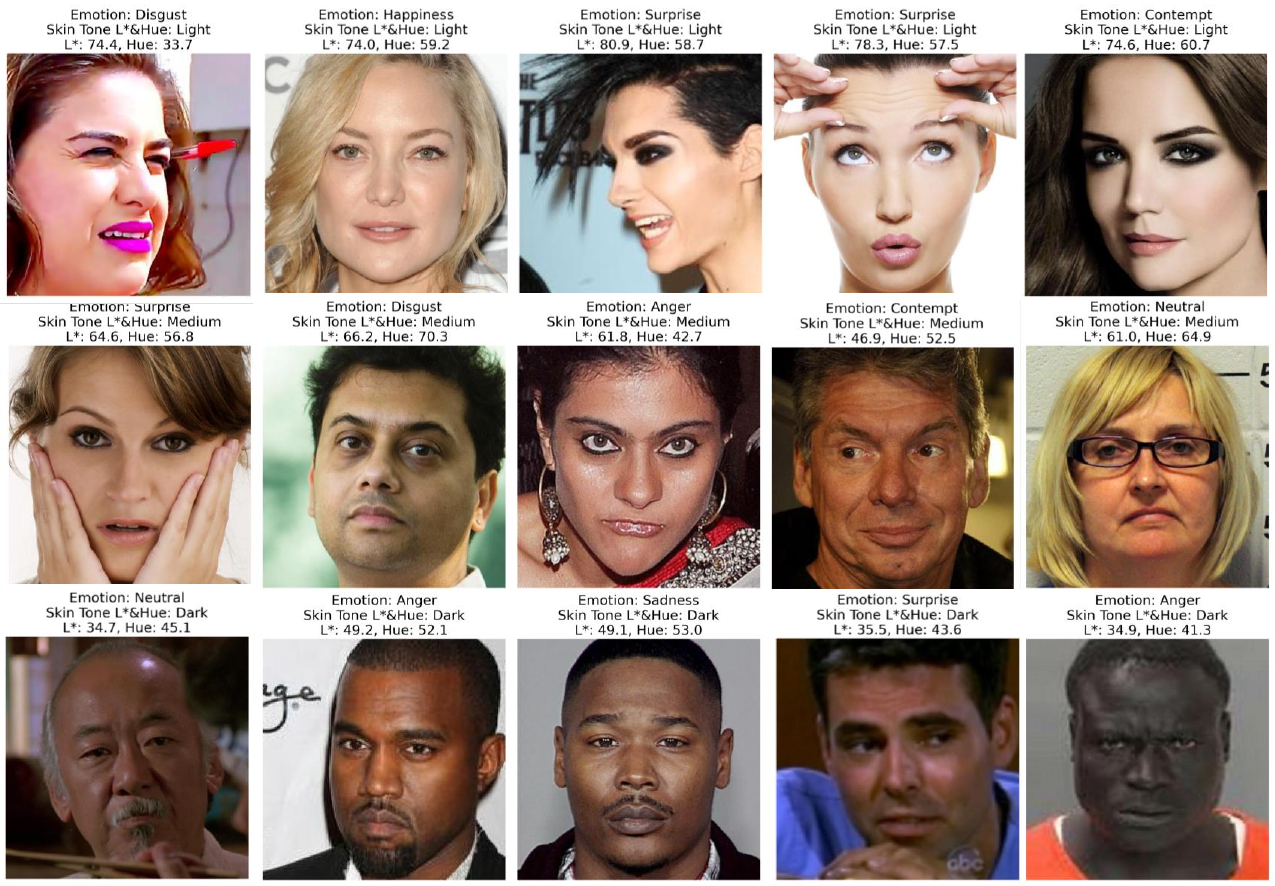}
    \caption{
    Proposed classification using \(L^*\) and Hue (\(H^* = \arctan(b^*/a^*)\)), capturing both lightness and chromaticity. Images were randomly sampled from the training subset of AffectNet. This method improves tone estimation, particularly under variable lighting and darker skin tones.
    }
    \label{fig:hue_method}
\end{figure}

As shown in Figures~\ref{fig:ita_method} and~\ref{fig:hue_method}, both methods generally agree on Light skin tones due to their shared reliance on \(L^*\). However, substantial differences appear in the classification of Medium and Dark tones. The ITA-based method tends to misclassify individuals with darker hues but higher \(L^*\) values, often influenced by illumination as ITA-Medium. In contrast, the proposed \(H^*-L^*\) method more reliably captures chromatic characteristics, particularly brown tones, assigning these cases to  Dark via its hue component and brown-tone override. In addition, its ability to filter out chromatically implausible samples improves classification quality and supports more accurate downstream fairness evaluations.

\subsection{Fairness Evaluation Framework}

To evaluate how facial affect recognition performance varies across skin tone groups, we adopt a multifaceted fairness evaluation framework grounded in established metrics from algorithmic fairness literature~\cite{Chiu2024,Bellamy2019,Gustafson2023,FAn2023}. Specifically, we compare group-level performance using the two described skin tone categorization methods.

We evaluate fairness using three primary group-based metrics:

\begin{itemize}
    \item \textbf{F1-score Gap:} The absolute difference in macro-averaged F1 score between the best and worst performing skin tone groups. This captures disparities in classification quality across demographic partitions.
    
    \item \textbf{Accuracy Equality:} The range in overall classification accuracy across Light, Medium, and Dark skin tone groups. A fair model should show minimal accuracy variance across these groups.
    
    \item \textbf{Equal Opportunity Difference (EOD):} Following~\cite{Hardt2016}, we define EOD as the disparity in class-wise recall (true positive rate) across skin tone groups. Specifically, for each emotion class \( c \) and group \( g \), recall is calculated as:
    \[
    \text{Recall}_{g,c} = \frac{\text{TP}_{g,c}}{\text{P}_{g,c}},
    \]
    where \( \text{TP}_{g,c} \) denotes the number of correctly predicted samples of class \( c \) in group \( g \), and \( \text{P}_{g,c} \) is the total number of ground truth samples of class \( c \) in that group. 
\end{itemize}

All fairness metrics are computed on the original unbalanced test subset. This design choice aligns with real-world deployment scenarios, where input distributions are not artificially manipulated to match ideal demographic proportions. In general, this framework supports both aggregate and per-class fairness evaluation, providing robust insights into the equity implications of training strategies.

\subsection{Explainability with Grad-CAM}

Explainable AI (XAI) is critical for improving transparency and interpretability in deep learning systems, especially in socially sensitive tasks such as FAA~\cite{Coroama2022}. Visual explanation methods help uncover how models interpret facial expressions across demographic groups, offering critical insights for fairness assessment ~\cite{biagi23, Mery2022,minplus}.

We adopt Gradient-weighted Class Activation Mapping (Grad-CAM)~\cite{Selvaraju2016}, a widely used technique that highlights the image regions most influential in predicting a model by tracing gradients back to the final convolutional layer. Grad-CAM produces class-specific heatmaps without altering the model architecture, making it well suited for CNN-based FAA systems. In our study, Grad-CAM is applied to representative samples from Light, Medium, and Dark skin tone groups (based on \(H^*-L^*\) space) to visualize attention patterns during emotion recognition. These overlays allow us to explore whether the model relies on consistent facial features across tones, supporting a qualitative evaluation of potential disparities in visual reasoning.

\section{Experiments and Results}

Figures~\ref{fig1} to~\ref{fig:hue_method} provide foundational results that inform our experimental design. They highlight key differences in the distribution and categorization of skin tones under the ITA and \(H^*\)-\(L^*\) methods, revealing substantial underrepresentation of darker tones and potential misclassification in the ITA approach. These findings motivate the fairness evaluations presented in this section, which apply both taxonomies to assess their impact on model performance across demographic groups.

\begin{figure}[ht]
    \centering
    \includegraphics[width=\linewidth]{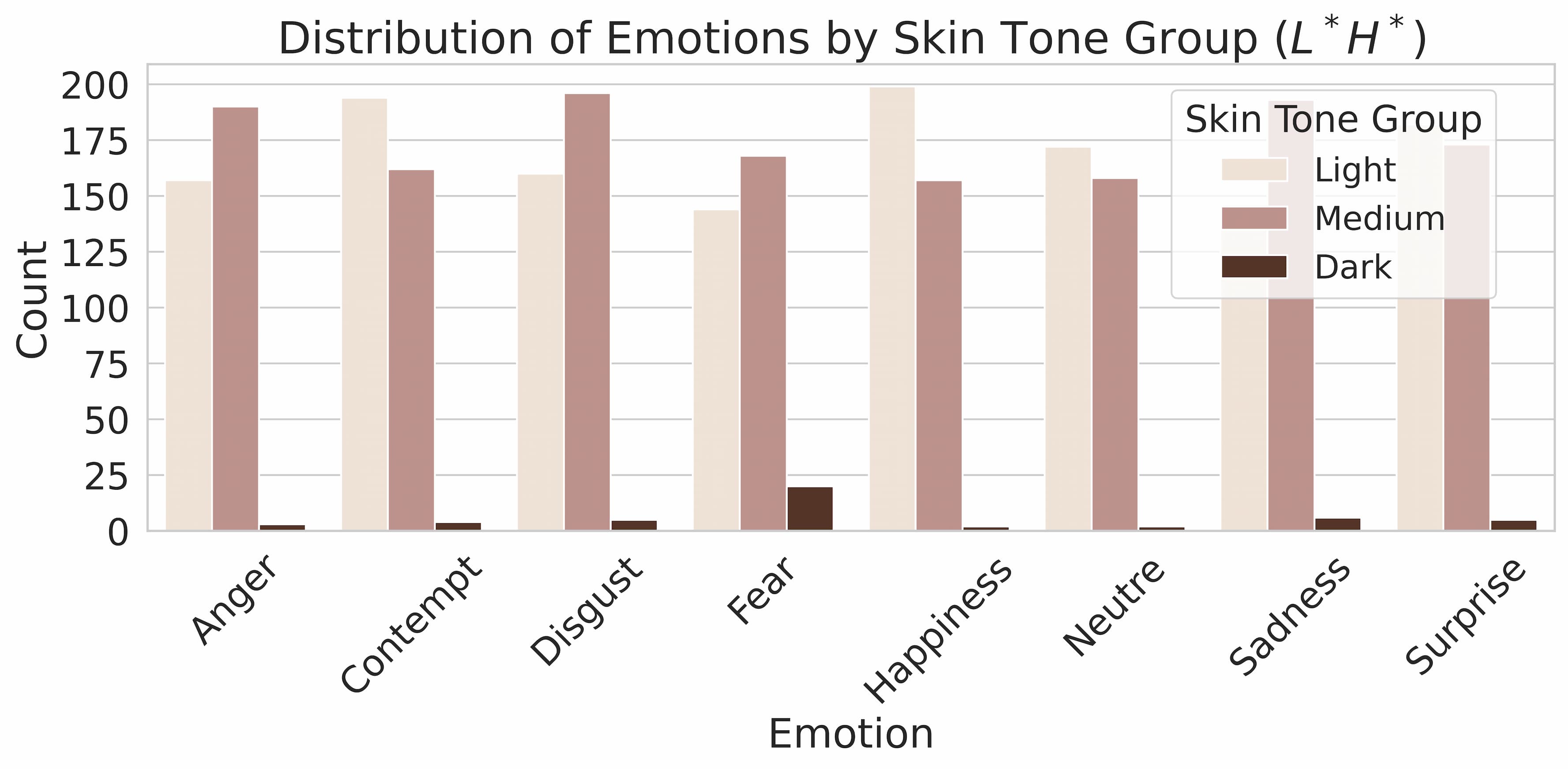}
    \caption{Distribution of labeled emotions in the AffectNet test set by skin tone groups classified using the \(H^*\)-\(L^*\) method.}
    \label{fig3}
\end{figure}

To establish a controlled evaluation framework, we employed a MobileNet-based classifier due to its low architectural complexity, widespread adoption, and suitability for real-time or embedded systems. This choice enables reproducibility and reduces confounding variables in fairness analyses, helping isolate the impact of skin tone group definitions on model performance. We evaluated the model on the unaltered AffectNet test set to reflect real-world demographic distributions and enable fairness auditing under realistic deployment conditions. As illustrated in Figure~\ref{fig3}, the distribution of emotion labels between skin tone groups is markedly imbalanced, with the dark group comprising only 52 of the 2,864 total images. Although this disparity limits the statistical power of group-specific analyses, particularly for the dark skin tone group, it reflects real-world demographic imbalances. We report disaggregated results to reveal the disparities that aggregated metrics can conceal. Future work should consider dataset balancing or re-weighting to improve subgroup robustness.

As shown in Table~\ref{tab0}, the baseline model demonstrates limited performance across most emotion categories, with macro-averaged precision, recall, and F1-score values around 0.40. We attribute these modest results to the model’s restricted representational capacity and to the strong class imbalance present in the AffectNet dataset. While suboptimal, this baseline can be used as a diagnostic tool. In the following sections, we apply both the traditional ITA-based grouping and our proposed \(H^*\)-\(L^*\)based classification to the baseline model, enabling a controlled comparison of fairness outcomes under each skin tone taxonomy.

\begin{table}[ht]
\centering
\caption{Per-emotion metrics (Precision, Recall, F1-score) of the baseline model}
\label{tab0}
\begin{tabular}{lccc}
\textbf{Emotion} & \textbf{Precision} & \textbf{Recall} & \textbf{F1-score} \\
\hline
\multicolumn{4}{l}{\textbf{Baseline}} \\
Anger     & 0.29 & 0.43 & 0.35 \\
Disgust   & 0.46 & 0.38 & 0.42 \\
Fear      & \textbf{0.62} & 0.37 & 0.46 \\
Happiness & 0.56 & \textbf{0.66} & \textbf{0.61} \\
Sadness   & 0.39 & 0.27 & 0.32 \\
Surprise  & 0.38 & 0.43 & 0.41 \\
Neutral   & 0.24 & 0.38 & 0.30 \\
Contempt  & 0.44 & 0.29 & 0.35 \\
\hline
Accuracy & 0.40 & 0.40 & 0.40 \\
Macro Avg & 0.42 & 0.40 & 0.40 \\
Weighted Avg & \textbf{0.43} & 0.40 & 0.40 \\
\end{tabular}
\end{table}

\subsection{Fairness Evaluation: ITA vs. \texorpdfstring{\(H^*\)-\(L^*\)}{L*H*} Groupings}

We evaluated classification performance across demographic subgroups using both the traditional ITA-based classification and our proposed \(H^*\)-\(L^*\)based method. Table~\ref{tab2}, \ref{tab3} and \ref{tab4} summarize the aggregate precision, recall, and F1-scores by skin tone group for each method. 

\begin{table}[ht]
\centering
\caption{Precision per emotion by skin tone group for ITA and Hue classification.}
\label{tab2}
\begin{tabular}{lccc}
\toprule
 & Light  & Medium  & Dark \\       
Emotion &  ITA / $H^*L^*$ &  ITA / $H^*L^*$ &  ITA / $H^*L^*$\\
\hline
Anger & 0.29 / \textbf{0.31} & 0.32 / \textbf{0.31} & 0.25 / \textbf{0.33} \\
Disgust & \textbf{0.47} / 0.41 & 0.46 / \textbf{0.50} & 0.67 / \textbf{1.00} \\
Fear & 0.62 / \textbf{0.63} & \textbf{0.53} / 0.50 & 0.50 / \textbf{0.78} \\
Happiness & 0.56 / \textbf{0.58} & \textbf{0.55} / 0.54 & \textbf{0.70} / 0.33 \\
Sadness & 0.39 / \textbf{0.42} & \textbf{0.42} / 0.40 & \textbf{0.60} / 0.29 \\
Surprise & 0.38 / \textbf{0.43} & \textbf{0.43} / 0.39 & \textbf{0.83} / 0.12 \\
Neutre & 0.24 / \textbf{0.31} & 0.28 / \textbf{0.23} & 0.13 / \textbf{0.25} \\
Contempt & 0.44 /\textbf{0.45} & \textbf{0.40} / 0.38 & \textbf{0.60} / 0.60 \\
\bottomrule
\end{tabular}
\end{table}

Table~\ref{tab2} reports precision per emotion across skin tone groups, comparing the ITA-based classification with the proposed \(H^*\)-\(L^*\) approach. Overall, performance varies notably by both emotion and tone group, with no method consistently outperforming the other. For Light skin tones, precision is largely comparable between methods, reflecting their shared reliance on \(L^*\) and the strong presence of lighter samples in the dataset. Greater differences emerge in the Medium group, where \(H^*\)-\(L^*\) shows lower precision in several emotions—e.g., Disgust (0.50 vs. 0.46), Fear (0.50 vs. 0.53), and Neutral (0.23 vs. 0.28). 

The most pronounced divergences appear in the Dark group. Here, \(H^*\)-\(L^*\) outperforms ITA in Disgust (1.00 vs. 0.67) and Fear (0.78 vs. 0.50), supporting the inclusion of hue for capturing chromatic nuances in darker complexions. Nevertheless, ITA exhibits superior precision for Happiness (0.70 vs. 0.33) and Surprise (0.83 vs. 0.12) within this group. It is important to note that, as demonstrated in the preceding section, a significant proportion of samples classified within the Dark group by ITA, in fact, correspond to the Medium or even Light groups. Consequently, the reliability of these ITA metrics may be compromised.

\begin{table}[ht]
\centering
\caption{Recall per emotion by skin tone group for ITA and Hue classification.}
\label{tab3}
\begin{tabular}{lccc}
\toprule
 & Light  & Medium  & Dark \\       
Emotion &  ITA / $H^*L^*$ &  ITA / $H^*L^*$ &  ITA / $H^*L^*$\\
\hline
Anger & \textbf{0.43} / 0.41 & 0.44 / 0.44 & 0.14 / \textbf{0.67} \\
Disgust & \textbf{0.38} / 0.35 & 0.43 / \textbf{0.45} & \textbf{0.50} / 0.20 \\
Fear & \textbf{0.37} / 0.33 & \textbf{0.19} / 0.27 & \textbf{0.40} / 0.35 \\
Happiness & 0.66 / \textbf{0.64} & 0.70 / \textbf{0.73} & \textbf{0.78} / 0.50 \\
Sadness & \textbf{0.27} / 0.29 & \textbf{0.22} / 0.22 & \textbf{0.60} / 0.33 \\
Surprise & 0.43 / \textbf{0.49} & \textbf{0.48} / 0.41 & \textbf{0.56} / 0.20 \\
Neutre & 0.38 / \textbf{0.47} & \textbf{0.40} / 0.32 & 0.67 / \textbf{1.00} \\
Contempt & 0.29 / \textbf{0.32} & \textbf{0.34} / 0.31 & 0.30 / \textbf{0.75} \\
\bottomrule
\end{tabular}
\end{table}

Table~\ref{tab3} shows recall for each emotion across skin tone groups. For Light and Medium skin, ITA and Hue perform comparably, with slight \(H^*\)-\(L^*\) advantages (e.g., Anger, Surprise, Neutre). Dark skin reveals key differences: Hue consistently exhibits substantially higher recall for most emotions (e.g., Anger, Disgust, Fear), indicating its superior ability to capture true positives in this group. Crucially, given ITA's tendency to misclassify Medium/Light skin as Dark, Hue's higher Dark skin recall likely reflects improved accuracy on actual Dark skin, while ITA's lower recall is partly due to incorrect skin tone assignments.

\begin{table}[ht]
\centering
\caption{F1-Score per emotion by skin tone group for ITA and Hue classification.}
\label{tab4}
\begin{tabular}{lccc}
\toprule
 & Light  & Medium  & Dark \\       
Emotion &  ITA / $H^*L^*$ &  ITA / $H^*L^*$ &  ITA / $H^*L^*$\\
\hline
Anger & 0.35 / \textbf{0.35} & 0.37 / \textbf{0.36} & 0.18 / \textbf{0.44} \\
Disgust & \textbf{0.42} / 0.38 & 0.44 / \textbf{0.48} & \textbf{0.57} / 0.33 \\
Fear & \textbf{0.46} / 0.43 & \textbf{0.28} / 0.35 & \textbf{0.44} / 0.48 \\
Happiness & 0.61 / \textbf{0.61} & 0.62 / \textbf{0.62} & \textbf{0.74} / 0.40 \\
Sadness & 0.32 / \textbf{0.34} & \textbf{0.29} / 0.28 & \textbf{0.60} / 0.31 \\
Surprise & \textbf{0.41} / 0.46 & \textbf{0.45} / 0.40 & \textbf{0.67} / 0.15 \\
Neutre & 0.30 / \textbf{0.37} & \textbf{0.33}/ 0.27 & 0.22 / \textbf{0.40} \\
Contempt & 0.35 / \textbf{0.37} & \textbf{0.37} / 0.34 & \textbf{0.40}/ \textbf{0.67} \\
\bottomrule
\end{tabular}
\end{table}

Finally, Table~\ref{tab4} presents the F1-score, balancing precision and recall, for each emotion across skin tone groups. Similar to previous observations, performance is generally comparable between ITA and \(H^*\)-\(L^*\) for Light and Medium skin. However, in Dark skin, \(H^*\)-\(L^*\) demonstrates a clear advantage for Disgust and Fear, confirming its improved ability to both accurately identify these emotions and capture most instances. Overall, F1-score results reinforce that \(H^*\)-\(L^*\) offers a more balanced performance in recognizing emotions within the Dark skin group, while the choice between methods is less critical for Light and Medium skin.

\section{Analysis of Fairness Disparity}

\begin{figure}[ht]
    \centering
    \includegraphics[width=\linewidth]{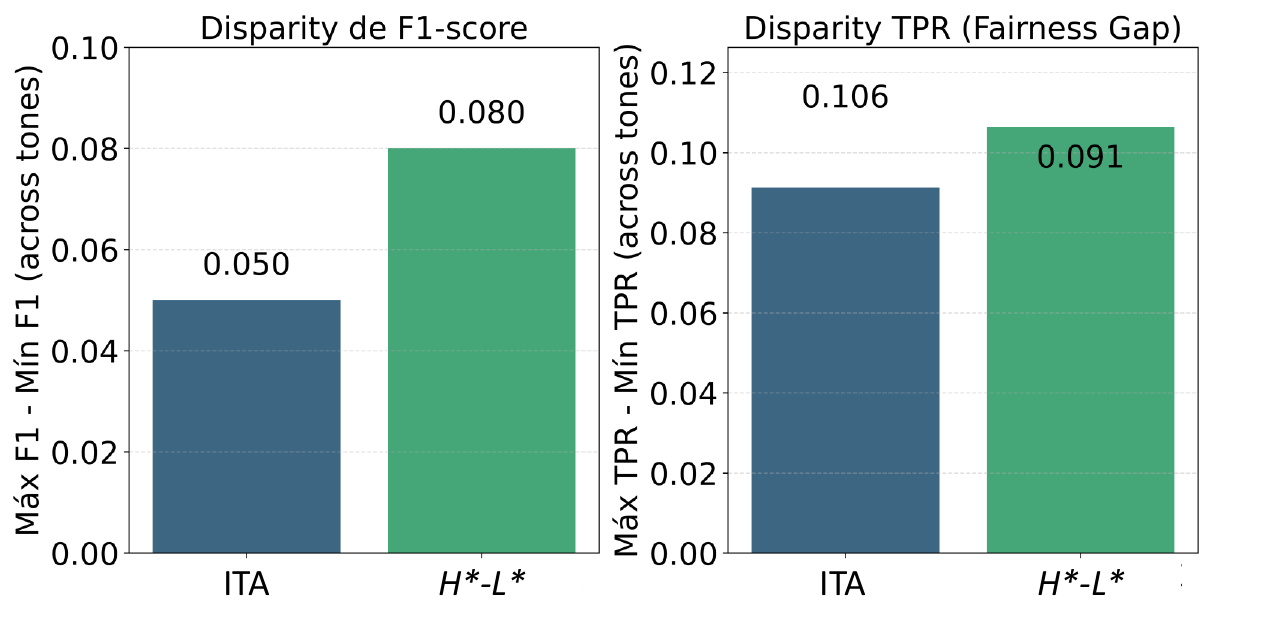}
    \caption{Comparison of F1-score and TPR disparity (fairness gap) between the ITA and \(H^*\)-\(L^*\) methods.}
    \label{fig6}
\end{figure}

Figure~\ref{fig6} presents a comparative analysis of F1-score and True Positive Rate (TPR) disparities across skin tone groups for the ITA and \(H^*\)-\(L^*\) skin tone classification methods, as evaluated within the baseline model. The F1-score disparity, depicted in the left panel, reveals that the \(H^*\)-\(L^*\) method exhibits greater performance variation across skin tones (disparity of 0.080) than the ITA method (disparity of 0.050). Similarly, the right panel illustrates the TPR disparity, where the \(H^*\)-\(L^*\) method again demonstrates a larger fairness gap (0.106) compared to the ITA method (0.091).

These results indicate that, while both methods exhibit some degree of performance inconsistency across skin tone groups, the \(H^*\)-\(L^*\) method demonstrates a more pronounced disparity in both F1-score and TPR. This suggests that the \(H^*\)-\(L^*\) method's performance demonstrates reduced equity, characterized by increased variability in accuracy and recall across different skin tones compared to the ITA method. Consequently, the ITA method, despite its inherent limitations, may offer marginally improved fairness in terms of overall performance consistency across the defined demographic subgroups.

\subsection{Equal Opportunity Analysis}

To further evaluate fairness, we examined the Equal Opportunity metric, specifically focusing on recall within each class (emotion) across skin tone groups. Equal Opportunity assesses whether the true positive rates are equal across different groups. In the context of emotion recognition, it measures if the model is equally capable of correctly identifying a specific emotion across all skin tone groups, given that the ground truth label is that emotion.

\begin{figure}[htbp]
  \centering
  \includegraphics[width=\linewidth]{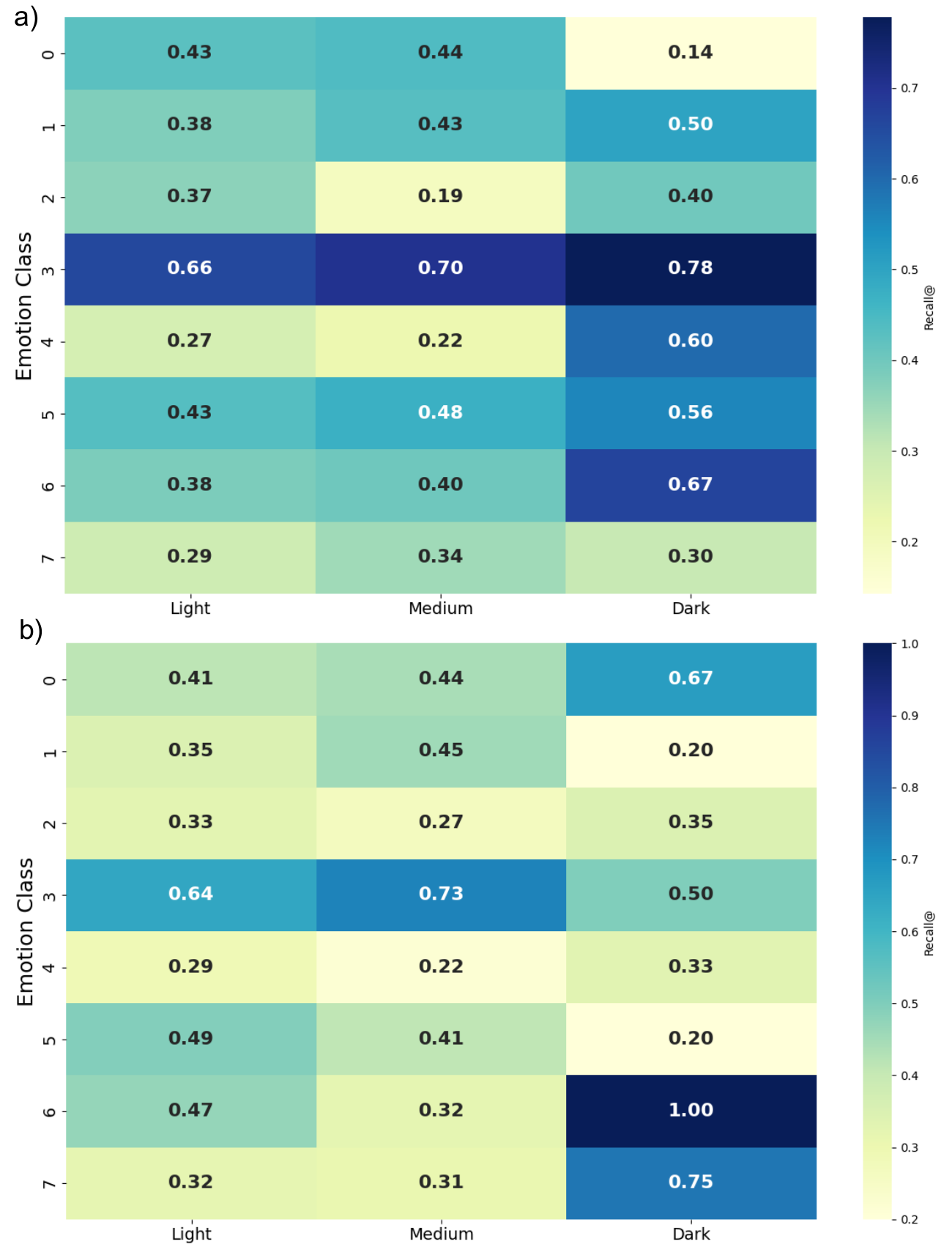}
  \caption{Equal Opportunity (Recall) for ITA (a) and \(H^*\)-\(L^*\) (b) across skin tone groups and emotions.}
  \label{FigRecall}
\end{figure}

Figure~\ref{FigRecall} (Panels a and b) presents a visual comparison of Equal Opportunity, specifically recall per emotion, across skin tone groups for the ITA (a) and \(H^*\)-\(L^*\) (b) methods.  The heatmaps provide a more granular view of the performance disparities observed in the previous aggregated TPR analysis and offer insights into potential biases in emotion recognition across skin tones.

For Light and Medium skin tones, the recall values for ITA and \(H^*\)-\(L^*\) are generally comparable. While there are some minor variations, neither method consistently demonstrates a clear advantage. This aligns with the observation of similar overall TPR for Light and Medium skin in Figure~\ref{fig6}.

However, the most notable differences emerge in the Dark skin group. \(H^*\)-\(L^*\) consistently exhibits substantially higher recall for most emotions (e.g., Anger, Disgust, Fear, Neutre, Contempt) compared to ITA. This improved recall for \(H^*\)-\(L^*\) in Dark skin is particularly relevant, considering the higher TPR disparity observed for \(H^*\)-\(L^*\) in Figure~\ref{fig6}. Despite having a larger overall fairness gap, \(H^*\)-\(L^*\)'s better recall on the Dark skin subgroup suggests it captures more true positive instances within that group, which is a crucial aspect of fairness.

Both ITA and \(H^*\)-\(L^*\) show relatively good performance in recognizing Happiness across all skin tone groups. The recall for Surprise tends to be more variable, especially for the Dark skin group, where both methods show lower recall compared to Light and Medium. This variability contributes to the overall TPR disparity seen in Figure~\ref{fig6}, indicating that both methods struggle more with Surprise in darker skin tones. For emotions like Anger, Disgust, Fear, Sadness, Neutre, and Contempt, recall values often vary between skin tone groups, with a trend towards lower recall for Dark skin, particularly for ITA. This pattern further exacerbates the fairness gap, as the models are less effective at recognizing these emotions in individuals with darker skin.

It is crucial to interpret the ITA results, especially for the Dark skin group, with caution. As discussed in Section~\ref{sec:methodology}, the ITA method has limitations in accurately classifying darker skin tones. Due to its reliance on lightness and omission of hue information, ITA may misclassify some individuals with darker skin as having Medium skin. This is visually supported by the heatmaps in Figure~\ref{FigRecall}, where the distribution of recall values for ITA in the Dark skin group appears more similar to the Medium group than in the \(H^*\)-\(L^*\) method, especially for emotions where \(H^*\)-\(L^*\) shows a significant improvement.

Consequently, the recall values for ITA in the Dark skin group might be artificially inflated. A higher recall for ITA in Dark skin could be partially attributed to the fact that it's incorrectly including some Medium-skinned individuals in that group. This means that ITA might appear to be good at identifying emotions in Dark skin, but it's actually looking at a mixed population.

Therefore, while ITA might show seemingly competitive or even better recall for certain emotions in the Dark skin group, this result needs to be considered in the context of its classification inaccuracies. The \(H^*\)-\(L^*\) method, designed to address these limitations, provides a more reliable assessment of emotion recognition performance across skin tones. This highlights a potential trade-off: while \(L^*H^*\) offers better subgroup fidelity, it may also make existing disparities more visible by capturing finer-grained differences.

\subsection{Explainability Results with Grad-CAM}

\begin{figure}[htbp]
    \centering
    \includegraphics[width=\linewidth]{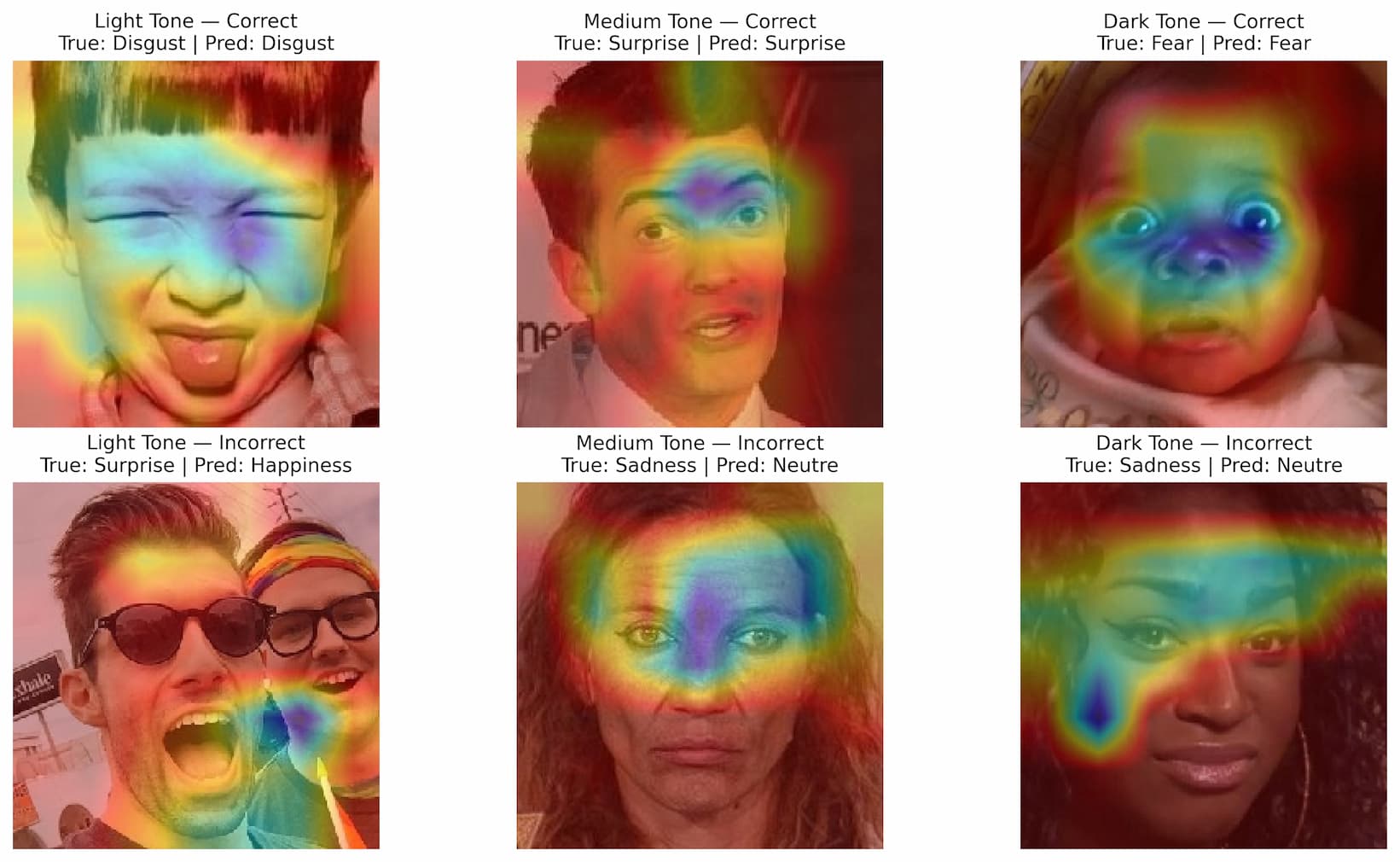}
    \caption{
    Grad-CAM visualizations of facial emotion predictions across skin tone groups using \(L^*H^*\)-based classification. 
    Each column corresponds to a different skin tone group—Light (left), Medium (center), and Dark (right)—with the top row showing correct predictions and the bottom row showing misclassified examples. 
    Heatmaps indicate the model's attention regions, with warmer colors reflecting greater influence on the final prediction. These visual diagnostics help illustrate differences in feature localization across skin tones.
    }
    \label{Fig7}
\end{figure}

To complement the quantitative fairness metrics, we used Grad-CAM to visualize the spatial attention patterns of the model across skin tone groups. Figure~\ref{Fig7} shows Grad-CAM overlays for representative samples from Light, Medium, and Dark skin tones, defined using the \(H^*\)-\(L^*\) classification.

In the Light group, correct predictions (e.g., Disgust) show attention focused around the mouth, while misclassifications (e.g., Surprise as Happiness) exhibit more diffuse activation.

In the Medium group, correct Surprise predictions highlight eye and brow regions, whereas misclassified Sadness samples show scattered attention.

In the Dark group, correct Fear predictions emphasize the eye region, but misclassified samples (e.g., Sadness) focus on less informative areas like the forehead and nose.

Overall, accurate predictions tend to correspond to localized, expression-relevant regions, while misclassifications show more dispersed or inconsistent patterns. However, the explainability findings remain limited in interpretability. As a post-hoc saliency method, Grad-CAM provides only coarse attribution and may not capture the true decision logic of the model—particularly across underrepresented groups. To strengthen the analysis, future work could integrate more robust explainability techniques, such as MinPlus \cite{minplus} or Layer-wise Relevance Propagation (LRP) \cite{bach2015,seibold2021}, which may offer deeper insight into model behavior and bias sources.

\begin{figure*}[t]
    \centering
    \includegraphics[width=0.9\linewidth]{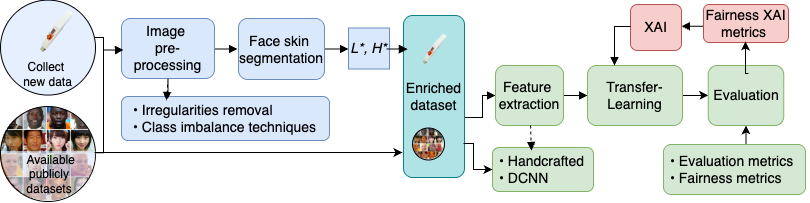}
    \caption{
        Overview of the proposed fairness-aware FAA pipeline. The process begins with a facial dataset that undergoes preprocessing and objective skin tone estimation using the \(H^*\)-\(L^*\) method. Images are then stratified by skin tone and resampled to mitigate representational imbalance. Emotion recognition models are trained via transfer learning and evaluated with fairness-aware metrics and explainability tools such as Grad-CAM.
    }
    \label{fig8}
\end{figure*}

\section{Overview of the Proposed Fairness-Aware Pipeline}

Figure~\ref{fig8} presents the architecture of the proposed fairness-aware Facial Affect Analysis pipeline, which integrates perceptual skin tone estimation, fairness evaluation, and model interpretability. The pipeline begins with data acquisition, combining publicly available datasets with the potential for curated collection using standardized measurement tools. In particular, future extensions may incorporate spectrophotometer-based measurements to establish a reliable reference standard for skin tone, enhancing the validity of dataset annotations and subsequent fairness evaluations.

The workflow includes image preprocessing, face skin segmentation, and objective skin tone classification using Lightness ($L^*$) and Hue ($H^*$). While this work focuses on comparative fairness assessment, the framework is designed to accommodate class balancing techniques to mitigate skin tone distribution biases. The enriched dataset feeds into a feature extraction stage supporting both handcrafted and deep learning-based representations followed by transfer learning and evaluation.

Model performance is assessed using conventional metrics (e.g., Accuracy, F1-score) alongside fairness indicators (e.g., TPR disparity, Equal Opportunity). Finally, explainability tools such as Grad-CAM are used to interpret model focus across demographic groups, supported by fairness-aware XAI metrics. This modular pipeline supports reproducible and extensible fairness assessments in FAA systems and can be adapted to other sensitive attributes.

\section{Toward Fair FAA: Challenges and Insights}
\label{sec6}

This work presented a fairness-aware pipeline for facial affect analysis leveraging a perceptually grounded skin tone classification method based on Lightness ($L^*$) and Hue ($H^*$). The proposed approach aims to address demographic disparities—particularly for underrepresented darker skin tones—by improving skin tone estimation and enabling more reliable fairness evaluations.

Our analysis, using fairness metrics such as F1-score disparity, TPR disparity, and Equal Opportunity, reveals that ITA and \(H^*\)-\(L^*\) lead to different subgrouping outcomes. While ITA shows slightly lower disparities in aggregate, its reliance solely on lightness and b* can result in misclassification—particularly by grouping Medium and Dark tones together—masking true performance gaps. In contrast, the \(H^*\)-\(L^*\) color space incorporates chromatic information, producing more consistent and discriminative subgroup definitions, especially for Dark skin tones. This allows for clearer identification of disparities in emotion recognition, notably in emotions such as \textit{Anger}, \textit{Fear}, and \textit{Contempt}. 

Explainability analysis using Grad-CAM further reveals that model attention varies by skin tone: focused and semantically meaningful for Light and Medium groups, but more diffuse or misaligned for Dark tones. This suggests a lack of feature robustness and highlights the importance of architectural transparency alongside data interventions.

To our knowledge, this is the first study to operationalize and compare the \(H^*\)-\(L^*\) based classification method in depth, a technique often cited but seldom implemented in fairness research. Our results suggest that prior studies relying exclusively on ITA may need to be interpreted with caution, as their skin tone groupings could underestimate disparities affecting darker-skinned individuals. While both methods are useful proxies in the absence of self-reported data, our results support \(H^*\)-\(L^*\) as a more perceptually grounded and robust alternative. Its rule-based corrections address ITA’s limitations, especially in underrepresented tone ranges.

Despite its advantages, the \(L^*H^*\)-based approach has limitations. It depends on empirically defined thresholds that may require tuning across datasets and remains sensitive to lighting and occlusion. 
Discretizing skin tone, a continuous attribute, introduces unavoidable simplifications. An additional consideration is that improved subgroup precision can sometimes amplify observed disparities by revealing biases masked by noisier proxies like ITA. Higher disparity, in this case, may reflect more accurate measurement rather than worsened fairness. While corrective steps like the brown-tone override help address edge cases, they too rely on dataset-specific heuristics. This highlights the importance of jointly optimizing group fidelity and outcome equity in fairness evaluation.

Future work will focus on using the \(H^*\)-\(L^*\) method to rebalance training data and evaluate fairness across deeper and more expressive model architectures. Ongoing experiments aim to determine how architectural complexity interacts with dataset composition in reducing bias. Specifically, we plan to extend the proposed fairness pipeline to convolutional models such as ResNet and transformer-based architectures like ViT, in order to assess whether increased capacity or self-attention mechanisms mitigate or exacerbate the skin tone disparities observed in lightweight CNNs.

In summary, improving fairness in FAA requires perceptually accurate skin tone estimation, representative data distributions, and explainability tools to uncover latent model biases. The \(H^*\)-\(L^*\) method provides a promising step in this direction, offering a more nuanced and reliable foundation for fairness auditing in facial analysis systems.

%%%%%%%%%%%%%%%%%%%%%%%%%%%%%%%%%%%%%%%%%%%%%%%

\section*{Ethical Impact Statement}

This work examines fairness in Facial Affect Analysis (FAA), focusing on skin tone disparities. Although no new data were collected and no human subjects were involved, FAA systems are increasingly deployed in sensitive domains such as mental health, education, and human-computer interaction, where biased output can lead to exclusion or loss of trust, especially for people with darker skin tones.

Our analysis highlights the underrepresentation of dark skin tones in AffectNet and reveals measurable fairness gaps in model performance. In response, we propose a perceptually grounded skin tone estimation method (\(H^*\)-\(L^*\)) and a fairness-aware evaluation pipeline that integrates explainability tools and diagnostic metrics.

While we do not apply dataset rebalancing in this work, our findings lay the groundwork for future mitigation strategies. We encourage the integration of perceptual grouping with data- and model-level fairness techniques and advocate for greater transparency and demographic representation in benchmark datasets.

This research supports ethical AI by exposing measurement bias, promoting reproducible fairness evaluations, and informing a more inclusive FAA system design.

\section*{Acknowledgment}

A.M.C. and D.M. acknowledge support from the National Center for Artificial Intelligence CENIA FB210017, Basal ANID, and ANID Millennium Science Initiative Program (iHealth) ICN2021\_004. A.M.C. also acknowledges support from ANID Project SA77210039 and Fondecyt Regular 1251081.

\small
\bibliographystyle{ieeetr}
\bibliography{FAA}

\end{document}